\documentclass{article}
\usepackage[T1]{fontenc}

\usepackage{microtype}
\usepackage{graphicx}
\usepackage{booktabs}
\usepackage{hyperref}

\usepackage[preprint]{icml2026}

\usepackage{amsmath}
\usepackage{amssymb}
\usepackage{mathtools}
\usepackage{amsthm}
\usepackage{multirow}
\usepackage{array}
\usepackage{xcolor}
\usepackage[capitalize,noabbrev]{cleveref}
\usepackage{algorithm}
\usepackage{algorithmic}

\theoremstyle{plain}
\newtheorem{lemma}{Lemma}
\newtheorem{proposition}{Proposition}
\theoremstyle{definition}
\newtheorem{definition}{Definition}

\crefname{lemma}{Lemma}{Lemmas}
\crefname{proposition}{Proposition}{Propositions}
\crefname{definition}{Definition}{Definitions}

\newcommand{\sd}[2]{#1\,$\pm$\,#2}
\DeclareMathOperator*{\argmin}{arg\,min}

\icmltitlerunning{Proximal Targets for Looped Refiners}

\begin{document}

\twocolumn[
  \icmltitle{Proximal Targets: Training Looped Refiners\\on Problems with Many Correct Answers}

  \begin{icmlauthorlist}
    \icmlauthor{Syed Affan Daimi}{iiit}
  \end{icmlauthorlist}

  \icmlaffiliation{iiit}{International Institute of Information Technology, Hyderabad, India}
  \icmlcorrespondingauthor{Syed Affan Daimi}{syed.daimi@research.iiit.ac.in}

  \icmlkeywords{looped transformers, recursive reasoning, deep supervision, multiple solutions, constraint satisfaction}

  \vskip 0.3in
]

\printAffiliationsAndNotice{}
% the style's running-title size test misfires under some engines (it does on its own example); set it directly
\makeatletter\gdef\@icmltitlerunning{Proximal Targets for Looped Refiners}\makeatother
\hypersetup{pdftitle={Proximal Targets: Training Looped Refiners on Problems with Many Correct Answers}}

\begin{abstract}
Looped refiners such as HRM and TRM apply one small network repeatedly to a draft answer, and deep supervision trains every pass toward one stored answer per puzzle.
We prove that when a puzzle has many correct answers this target sets a loss floor that no model capacity removes, and that the loss-minimising pass blends the answers into an invalid grid.
We propose the \emph{proximal target}: train each pass toward the correct answer nearest to its own input, found by scanning a list of answers, by a search solver started at the draft, or within a band of quality set by a knob; correct answers become fixed points, and a pass only has to make fewer errors than its input has for the loop to converge, which its training loss bounds.
With the same 1.7M-parameter network and 5 seeds, the proximal target solves 98.8\% of Sudoku puzzles with 2--64 answers (standard target: 2.5\%), 93.8\% of sparse puzzles with more than 10,000 answers without any list of answers (0.0\%), and 90.9\% of 200-node 3-colourings after training on graphs of at most 50 nodes (36.9\%), and different random starts end at different answers.
Against GRAM, a stochastic-latent reasoner, a single run is correct far more often (98.8\% vs.\ 57.7\%) and twenty runs find more of the answers (53.6\% vs.\ 37.4\%), while GRAM is slightly ahead on its own 10-node colouring benchmark; in Scrabble the knob becomes a calibrated strength dial.
\end{abstract}

%=====================================================================
\section{Introduction}
\label{sec:intro}

A small network applied over and over can solve puzzles that defeat much larger models.
The Hierarchical Reasoning Model (HRM;~\citealp{wang2025hrm}) and the Tiny Recursive Model (TRM;~\citealp{jolicoeur2025trm}) keep a \emph{draft} of the answer and apply one shared network to it repeatedly; each application, a \emph{pass}, reads the puzzle and the current draft and writes a better draft.
Training uses deep supervision: every pass is scored against the answer.

This works when the answer is unique.
Many problems we would like such refiners to solve are not like that: an under-constrained Sudoku, a graph colouring, a schedule, a move in a game.
Deep supervision then has to pick \emph{which} answer to train toward, and the standard choice is the one answer the dataset happens to store.
The choice looks harmless, but it is not.

\begin{figure*}[t]
  \centering
  \includegraphics[width=0.86\textwidth]{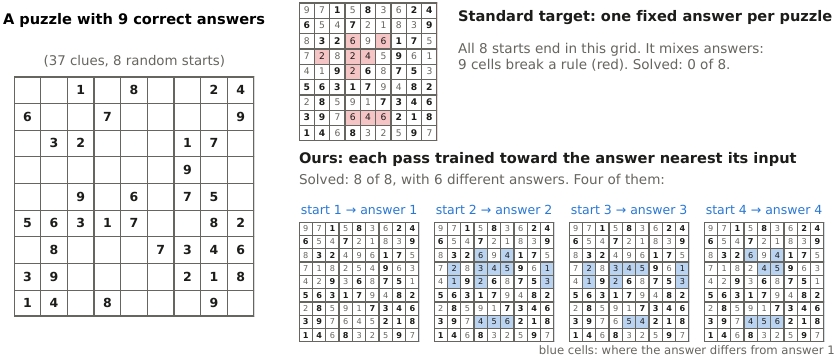}
  \caption{\textbf{Same network, same data, different target: training each pass toward the nearest correct answer removes blending.}
  Left: our running example, a Sudoku with 37 clues and 9 correct answers.
  Top: trained with the standard target (one stored answer per puzzle), the refiner sends all 8 random starts to the same grid, which mixes answers; 9 cells break a rule (red), and 0 of 8 starts solve.
  Bottom: trained toward the correct answer nearest each pass's input, all 8 starts solve and reach 6 different answers (four shown; blue marks cells that differ from answer~1).
  All grids are real outputs of the two trained models.}
  \label{fig:teaser}
\end{figure*}

A two-cell example shows why.
Suppose a puzzle has the two answers $(1,2)$ and $(2,1)$, and the dataset stores one of them for reasons the network cannot see.
The prediction that minimises the training loss puts 50\% on each digit in each cell, whatever the draft says.
Decoding it gives $(1,1)$ or $(2,2)$, neither of which is an answer.
The network has not failed to learn; it has learned the best response to a target that is arbitrary from where it stands.
On real puzzles this \emph{blending} is severe (\cref{fig:teaser}): a TRM-style refiner solves 96\% of Sudoku puzzles with one answer but 2.5\% of those with 2--64 answers, and making the network 4$\times$ larger does not help (2.4\%).

We change one thing: what each pass is trained toward.
The \emph{proximal target} is a correct answer chosen \emph{from the pass's own input}; in its simplest form, the correct answer closest to the current draft.
Because the target is now a function of what the network sees, there is nothing to blend.
A draft that is already correct is its own target, so every answer is a stable fixed point, and different random starts are pulled to different answers.
A pass does not even need to solve the puzzle: it only has to leave fewer wrong cells than it received, and the loop does the rest.

The target can come from a list of answers, from a depth-first solver started at the draft (no list needed), or, when answers differ in quality, from a quality band chosen by a knob.

\paragraph{Contributions.}
\begin{itemize}\setlength\itemsep{1pt}
  \item \textbf{A diagnosis.} When the stored answer cannot be predicted from the input, the loss-minimising pass outputs per-cell marginals over the answers and pays an irreducible loss (\cref{lem:blend}); trained models sit above this floor at every size we tried.
  \item \textbf{A fix with guarantees.} The proximal target makes every answer a fixed point, keeps the target unchanged while a draft is repaired, and turns a pass's training loss into a bound on the loop's progress (\cref{lem:fixed,lem:stay,prop:progress}); the lemmas are verified exhaustively on 40.3M small-puzzle drafts, and every statement is measured inside the trained loops.
  \item \textbf{Three target sources}, including a warm-started solver that needs no answer list, and a knob that controls the quality of the answer.
  \item \textbf{Evidence on four tasks with 5 seeds}, with controls and comparisons to MinLoss~\citep{nandwani2021oneofmany}, GRAM~\citep{baek2026gram} (official code), PTRM-style noise~\citep{sghaier2026ptrm} and ConsFormer~\citep{xu2025consformer}.
\end{itemize}

%=====================================================================
\section{Why the standard target blends}
\label{sec:problem}

A puzzle $x$ has a set $F$ of $m$ free cells and a set $S(x)$ of correct answers.
A \emph{draft} $s$ fills every free cell, and $d(s,y)$ counts the free cells in which $s$ and $y$ differ.
A pass reads the puzzle, the draft $s_{t-1}$ and a latent scratchpad $z_{t-1}$, and outputs a distribution $p_t$ over values for every free cell together with a new scratchpad; the most likely value in each cell gives the next draft $s_t$.
The refiner starts from a random draft $s_0\sim\mu$ and applies the same pass $N=8$ times.
Deep supervision scores every pass against a target $y_t$:
\begin{equation}
  \mathcal{L} \;=\; \sum_{t=1}^{N}\,\sum_{i\in F} -\log p_{t,i}\big(y_{t,i}\big).
  \label{eq:loss}
\end{equation}
The \emph{standard target} is one stored answer per training puzzle, the same for every pass and every draft.

\begin{lemma}[blending, and a floor]
\label{lem:blend}
Suppose that, given everything the pass sees, the target is drawn independently of the draft from a distribution $q_x$ over $S(x)$.
Then the pass that minimises~\eqref{eq:loss} outputs in every cell the marginal $p^\star_{t,i}(v)=q_x(y_i{=}v)$, and its expected loss at every pass is at least
$F(x)=\sum_{i\in F} H_{q_x}(y_i)$, whatever its capacity.
\end{lemma}

The premise concerns prediction, not storage.
The stored answer is a fixed function of the puzzle, but which answer a generator happened to store cannot be inferred from the clues, so on unseen puzzles $q_x$ is spread over $S(x)$.
The optimal pass therefore ignores the draft, splits probability wherever answers disagree, and decodes to a mixture; on our Sudoku test set the floor is $F=35.5$ bits per multi-answer puzzle.
Proofs of all statements are in \cref{app:proofs}.

\Cref{lem:blend} rules out the obvious repairs: more capacity does not lower $F$, and resampling the stored answer at every pass keeps it independent of the draft.
A target predictable from the puzzle, such as a fixed solver's first answer, has $F=0$, but asks a converged loop to move far (\cref{prop:disp}) and is hard to learn (\cref{sec:res-theory}).

%=====================================================================
\section{Proximal targets}
\label{sec:method}

\begin{definition}[proximal target]
For a cost $c: S(x)\to\mathbb{R}\cup\{+\infty\}$, the proximal target of a draft $s$ is
\begin{equation}
  T_c(s) \;=\; \argmin_{y\in S(x)}\; \big[\, d(s,y) + c(y) \,\big],
  \label{eq:proximal}
\end{equation}
with ties broken by a fixed order over answers.
\end{definition}

In words: the correct answer closest to the draft, after a penalty $c$ for less preferred answers.
With $c=0$ it is the \emph{nearest-answer target}.
Pass $t$ is trained toward $T_c(s_{t-1})$, computed on the drafts the model itself produces (\cref{alg:train}).
The target is now a function of the pass's input, so the floor of \cref{lem:blend} is zero: over all functions, the minimiser of~\eqref{eq:loss} outputs $T_c$ itself.

\begin{algorithm}[t]
  \caption{Deep supervision with a proximal target}
  \label{alg:train}
  \begin{algorithmic}
    \STATE {\bfseries Input:} puzzle $x$; target oracle $T$ (list scan, warm-started solver, or knob band); passes $N$
    \STATE $s_0 \sim \mu$ (random draft), $z_0 \leftarrow z_{\text{init}}$
    \FOR{$t=1$ {\bfseries to} $N$}
      \STATE $y_t \leftarrow T(x, s_{t-1})$ \hfill $\triangleright$ chosen from the input
      \STATE $(p_t, z_t) \leftarrow f_\theta(x, s_{t-1}, z_{t-1})$
      \STATE take a gradient step on $\sum_{i\in F} -\log p_{t,i}(y_{t,i})$
      \STATE $s_t \leftarrow \arg\max p_t$; \ detach $z_t$
    \ENDFOR
  \end{algorithmic}
\end{algorithm}

\subsection{What the target guarantees}
\label{sec:theory}

\begin{lemma}[every answer is a fixed point]
\label{lem:fixed}
$T_c(T_c(s)) = T_c(s)$ for every draft $s$.
With $c=0$ every correct answer is a fixed point; with $c\in\{0,+\infty\}$ the fixed points are exactly the answers with $c=0$.
\end{lemma}

A draft that is already correct is therefore trained to stay put, and a loop that follows its target from a random start $s_0$ ends at answer $y$ with probability $\mu(T_c^{-1}(y))$: the loop itself is deterministic, and diversity comes from where it starts.

\begin{lemma}[the target stays put during repair]
\label{lem:stay}
Let $y=T_c(s)$. If $s'$ sets any subset of the cells where $s$ differs from $y$ to $y$'s values, then $T_c(s')=y$.
\end{lemma}

Partial repairs made by successive passes therefore add up rather than chase a moving target.
The fixed tie order matters: with random tie-breaking, 1.8--3.3\% of multi-cell repairs on small puzzles change the target.
Both lemmas hold with zero exceptions on every draft of 60 $4{\times}4$ puzzles (40.3M drafts, 250.5M single-cell repairs).

A pass need not compute $T_c$, which is a search problem.
Let $V(s)=\min_y [d(s,y)+c(y)]$, the distance from a draft to the nearest (cost-adjusted) answer; with $c=0$ it counts the cells still to fix.

\begin{proposition}[progress without solving]
\label{prop:progress}
Let $s'$ be the decoded output of a pass on draft $s$, and $e=d(s',T_c(s))$ its errors against its own target.
Then $V(s')\le e + c(T_c(s))$; and under argmax decoding, $e\le \sum_{i\in F}-\log_2 p_i(T_c(s)_i)$, the pass's loss in bits.
\end{proposition}

So the loop makes progress whenever a pass leaves fewer errors than its input had ($e<V(s)$ when $c=0$); the training loss bounds those errors on every draft, not only on average; and if $e\le(1-\gamma)V(s)$ on the drafts the loop visits, $V$ falls as $(1-\gamma)^t$.
Whether $e<V$ holds is empirical; we measure it.
Training on the model's own drafts toward a correction of each draft is DAgger~\citep{ross2011dagger} with a projection as the expert, so training visits the states seen at test time.

\begin{proposition}[what a fixed target demands]
\label{prop:disp}
Let a target $\kappa(x)\in S(x)$ be fixed per puzzle, and $\delta(x)$ the least distance between two correct answers.
A pass whose input is a correct answer $y\ne\kappa(x)$ is trained to change at least $\delta(x)$ cells; under the nearest-answer target it is trained to change none.
\end{proposition}

For Sudoku $\delta\ge 4$.
A fixed but predictable target has no floor in principle, yet keeps asking a loop that has already converged to jump to another answer.

\subsection{Three sources of the target}
\label{sec:sources}

\paragraph{An answer list.}
When all correct answers can be listed, the nearest-answer target is a scan over the list.
The list must be complete: if only some answers are listed, the nearest \emph{listed} answer depends on which ones were kept, which the network cannot see, and \cref{lem:blend} returns in part (\cref{sec:res-lists}).

\paragraph{A search solver started at the draft.}
When answers are too many to list, let $\rho(s)$ be the first answer found by depth-first search that picks the most constrained cell first and tries the draft's value first in every cell.
This \emph{solver target} is not of the form~\eqref{eq:proximal}, but it has the two properties the loop uses: $\rho(y)=y$ for every correct $y$, and $\rho$ is unchanged when cells are repaired toward $\rho(s)$ (\cref{app:proofs}).
It costs one solver call per draft (55--75 search nodes on Sudoku) and needs no list.

\paragraph{A ranking and a knob.}
When answers differ in quality, the network receives an extra input $k\in[0,1]$, and $r(y)\in[0,1]$ is the quality rank of answer $y$.
The knob target is $T_k(s)=\argmin_y d(s,y)+c_k(y)$ with $c_k(y)=0$ if $|r(y)-k|\le w/2$ and $+\infty$ otherwise, so by \cref{lem:fixed} its fixed points are exactly the answers of the requested quality.
The knob decides \emph{which answers are stable}; temperature or latent noise only changes \emph{which basin a run falls into}.

\begin{figure*}[t]
  \centering
  \includegraphics[width=0.84\textwidth]{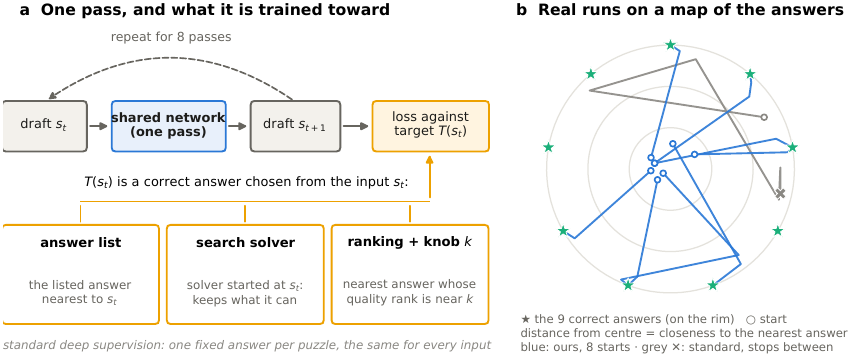}
  \caption{\textbf{Each pass is trained toward a correct answer chosen from its own input, so different starts travel to different answers.}
  (a)~A pass reads the draft and writes a new one, 8 times, carrying a scratchpad. Its target is the nearest listed answer, the answer a solver started at the draft finds, or (with a knob $k$) the nearest answer of quality near $k$. The standard target ignores the input.
  (b)~Real runs on the running example, drawn on a map of cell distances. Stars on the rim: the 9 correct answers; a draft's distance from the centre shows how close it is to its nearest answer.
  Our 8 runs start near the centre ($\circ$) and travel to different answers; the standard model's run stops between answers ($\times$).}
  \label{fig:method}
\end{figure*}

%=====================================================================
\section{Experimental setup}
\label{sec:setup}

\paragraph{Tasks.}
\emph{Sudoku with 2--64 answers}: $9{\times}9$ puzzles, half with a unique answer and half with 2--64 answers, all enumerated by an exact solver; the test set has 2,069 single-answer and 1,931 multi-answer puzzles, none sharing a solution grid with training even after relabelling digits.
\emph{Sparse Sudoku}: 17--28 clues and more than 10,000 answers, too many to list; targets come from the solver.
\emph{The multi-solution Sudoku benchmark} of \citet{nandwani2021oneofmany}, which lists at most 5 answers per training puzzle.
\emph{Planted 3-colouring}: random graphs of mean degree 4 with 15\% of nodes pre-coloured; training graphs have 20--50 nodes, test graphs up to 200; targets come from the solver.
\emph{Scrabble} on an $11{\times}11$ board, where MAGPIE~\citep{castellano2024magpie} labels every legal move with its win probability.

\paragraph{Model and training.}
All puzzle models use the TRM architecture~\citep{jolicoeur2025trm}: one token per cell, two transformer blocks of width 256, and two latent states carried between passes as the scratchpad (1.73M parameters for Sudoku, 1.71M for colouring, with attention along graph edges).
They run 8 passes in training and at test time from a uniform random start; the standard target uses TRM's blank start, and we also train it from random starts.
Every configuration sees the same 3M training examples.
Headline runs use 5 seeds and all others 3; we report mean $\pm$ standard deviation over seeds (\cref{app:details}).

\paragraph{Measures.}
A puzzle is \emph{solved} if the draft after 8 passes from one random start is correct.
\emph{Answers found} is the share of a puzzle's answers reached by 20 random starts, for puzzles with at most 64 answers (20 independent uniform draws over the answers would give 68.9\%).
Where answers cannot be listed we count \emph{distinct answers} among 20 starts.

%=====================================================================
\section{Results}
\label{sec:results}

\subsection{The proximal target solves where the standard target fails}
\label{sec:res-main}

\begin{figure*}[t]
  \centering
  \includegraphics[width=0.8\textwidth]{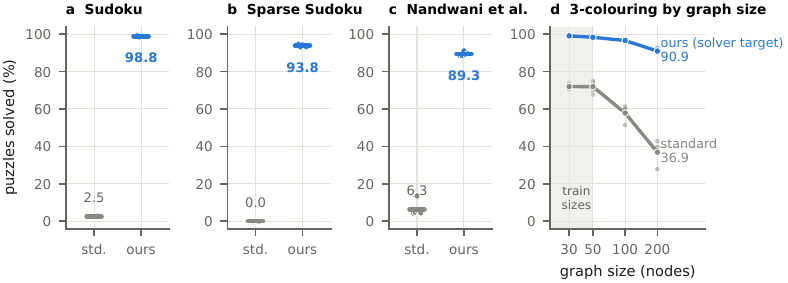}
  \caption{\textbf{The proximal target solves where the standard target fails, on every task.}
  Puzzles solved; dots are 5 training seeds, bars their mean.
  (a)~Sudoku with 2--64 answers. (b)~Sparse Sudoku with more than 10,000 answers, trained with the solver target and no answer list. (c)~The benchmark of \citet{nandwani2021oneofmany}. (d)~Planted 3-colouring trained on 20--50 nodes (shaded) and tested up to 200: the gap widens with size.}
  \label{fig:results}
\end{figure*}

\Cref{fig:results} shows the main result.
On Sudoku with 2--64 answers, the proximal target solves \sd{98.8}{0.3}\% of multi-answer puzzles against \sd{2.5}{0.2}\% for the standard target.
Without any answer list, the solver target solves \sd{93.8}{0.7}\% of sparse puzzles (standard: 0.0\%).
On the benchmark of \citeauthor{nandwani2021oneofmany} it solves \sd{89.3}{1.1}\% (\sd{6.3}{4.0}\%).
On colouring the standard target degrades as graphs grow, from 71.9\% at 30 nodes to \sd{36.9}{5.6}\% at 200, while the solver target stays at \sd{90.9}{1.0}\% on graphs four times larger than any seen in training.

The runs also spread over the answers: 20 starts find 53.6\% of a Sudoku puzzle's answers (standard: 0.8\%), 49.4\% on the benchmark (2.3\%), 15.1 distinct sparse-Sudoku answers, and 9.2 distinct 200-node colourings (0.9).

\paragraph{Controls.}
The standard target's failure is not caused by the blank start (from random starts: \sd{4.3}{2.9}\%), by too little supervision (a fresh random answer at every pass: \sd{3.2}{1.5}\%), or by capacity (4$\times$ parameters, 6.9M: \sd{2.4}{0.1}\%).
Conversely, the proximal target does not need random starts to solve (blank start: \sd{98.8}{0.2}\%, though it then finds fewer answers, 10.6\%) and keeps working at 4$\times$ size (\sd{99.1}{0.1}\%, answers found 55.0\%).

\begin{table*}[t]
  \caption{\textbf{Methods for many-answer puzzles on Sudoku with 2--64 answers} (1,931 multi-answer test puzzles; mean $\pm$ sd over seeds).
  Solved: one run is correct. Found: share of each puzzle's answers reached by 20 runs. ``Passes'' are applications of the network at test time (GRAM: adaptive steps of a nested recursion).
  $^\dagger$Share of noisy runs that are valid; with 20 runs, 24.4\% of puzzles have at least one.}
  \label{tab:main}
  \centering
  \small
  \setlength{\tabcolsep}{3.5pt}
  \begin{tabular}{@{}llrrrr@{}}
    \toprule
    Method & Each pass is trained toward \ldots & Params & Passes & Solved (\%) & Found (\%) \\
    \midrule
    Standard target (TRM) & one stored answer & 1.73M & 8 & \sd{2.5}{0.2} & \sd{0.8}{0.0} \\
    \quad 4$\times$ parameters & one stored answer & 6.9M & 8 & \sd{2.4}{0.1} & \sd{0.8}{0.0} \\
    \quad + test-time noise (PTRM-style) & one stored answer & 1.73M & 8 & 4.5$^\dagger$ & 8.8 \\
    Solver's first answer, fixed per puzzle & one predictable answer & 1.73M & 8 & \sd{51.3}{0.7} & \sd{8.3}{0.1} \\
    MinLoss~\citep{nandwani2021oneofmany} & answer nearest the \emph{output} & 1.73M & 8 & \sd{98.3}{0.3} & \sd{14.7}{0.2} \\
    GRAM~\citep{baek2026gram}, KL 0.045 & all answers, stochastic latent & 9.1M & 16 & \sd{57.7}{5.7} & \sd{37.4}{1.9} \\
    GRAM, KL 0.1 & all answers, stochastic latent & 9.1M & 16 & \sd{44.6}{16.1} & \sd{12.6}{1.7} \\
    ConsFormer~\citep{xu2025consformer} & fewer rule violations (no answers) & 1.84M & 8 & \sd{0.1}{0.1} & -- \\
    \quad iterated 1,000 times & & 1.84M & 1,000 & \sd{92.5}{0.5} & \sd{64.3}{0.3} \\
    \midrule
    \textbf{Proximal}: solver (no list) & answer a solver finds from the \emph{input} & 1.73M & 8 & \sd{97.0}{0.2} & \sd{31.5}{1.3} \\
    \textbf{Proximal}: nearest answer & answer nearest the \emph{input} & 1.73M & 8 & \textbf{\sd{98.8}{0.3}} & \sd{53.6}{0.4} \\
    \bottomrule
  \end{tabular}
\end{table*}

\subsection{Comparison with other approaches}
\label{sec:res-baselines}

\Cref{tab:main} places the proximal target among the approaches that address many answers, on the one task where every answer is known.

\begin{figure}[t]
  \centering
  \includegraphics[width=\columnwidth]{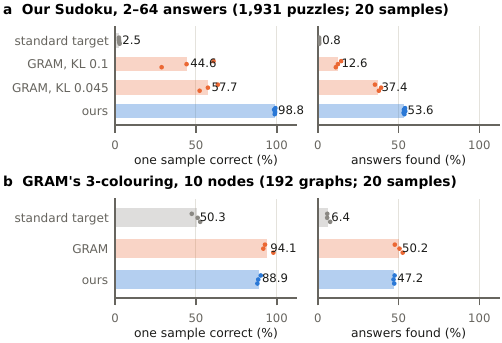}
  \caption{\textbf{GRAM, run from its official code, against the proximal target} (3 seeds each; dots are seeds).
  (a)~On Sudoku with 2--64 answers, GRAM's samples are diverse but a single one is often wrong.
  (b)~On GRAM's own 10-node colouring benchmark, where our runs of its code reproduce its published result, GRAM is slightly ahead.
}
  \label{fig:gram}
\end{figure}

\paragraph{GRAM.}
GRAM~\citep{baek2026gram} adds stochastic latent variables to a recursive reasoner and trains them by variational inference on every listed answer.
We ran its official code, unmodified, with its Sudoku configuration, a KL weight selected on our validation set, and a matched training budget (\cref{app:gram}); our runs reproduce its published result on its own colouring benchmark (50.0\% of colourings found; published 51.3\%).
Its samples are diverse but often wrong (\cref{fig:gram}a): one sample is correct for \sd{57.7}{5.7}\% of puzzles and the one its Q head picks for 65.7\%, although some sample among 20 is correct for 96.6\%.
On that benchmark GRAM is slightly ahead of our solver target (\cref{fig:gram}b): it finds 50.2\% of the colourings (up to relabelling of colours) against our 47.2\%, and a single sample is valid for 94.1\% of graphs against our 88.9\%; the standard target, trained on the same graphs, reaches 6.4\% and 50.3\%.

\paragraph{MinLoss, noise and rule penalties.}
Choosing the target from the pass's \emph{output}, as MinLoss does, solves but collapses onto a few answers (14.7\% found).
PTRM-style latent noise on the standard model raises the share of puzzles with any valid run among 20 from 2.3\% to 24.4\% at best: noise moves runs between basins, but the standard model's basins are not answers.
ConsFormer~\citep{xu2025consformer} trains one improvement step against differentiable rule penalties and is designed to be iterated hundreds to thousands of times: it solves 0.1\% at 8 passes, 92.5\% at 1,000 and 97.5\% at 2,000, and at 1,000 passes it finds more Sudoku answers than our 8-pass loop (64.3\% vs.\ 53.6\%).
Trained over 8 unrolled passes it still solves at most 0.2\%; the two work in different regimes, and only ConsFormer needs the rules as a formula.

\subsection{The theory, measured in the trained loops}
\label{sec:res-theory}

\begin{figure*}[t]
  \centering
  \includegraphics[width=0.8\textwidth]{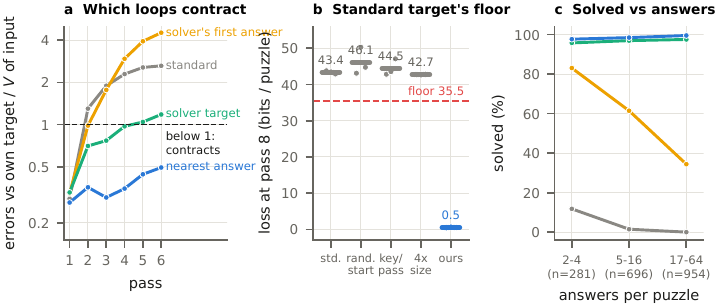}
  \caption{\textbf{The theory holds in the trained loops.}
  (a)~A pass's errors against its own target, divided by its input's distance to the nearest answer (mean of 3 seeds). Below 1, \cref{prop:progress} guarantees progress: the nearest-answer loop stays below 1, fixed targets cross it by pass 3. (The solver target is not the nearest answer, so its ratio near 1 is a loose bound, not a stall; it solves 97\%.)
  (b)~Loss at pass 8 of standard-target models trained from blank or random starts, with a new random answer at every pass, or at 4$\times$ size; all sit above the floor $F$ of \cref{lem:blend}.
  (c)~Solved by number of answers per puzzle: fixed targets fail more as answers multiply; targets chosen from the input do not.}
  \label{fig:theory}
\end{figure*}

\Cref{fig:theory} tests \cref{sec:problem,sec:theory} on 1,000 multi-answer test puzzles (4 starts each, 3 seeds).
\emph{The floor.} Every standard-target model ends above the floor of \cref{lem:blend} (42.7--46.1 bits against 35.5), whatever its start, supervision or size; ours reaches 0.5 bits.
\emph{Progress.} The two inequalities of \cref{prop:progress} hold in every run of all 17 models measured.
The nearest-answer loop leaves 0.28--0.49 times as many errors as its input had in passes 1--6, so it contracts; the standard and fixed-solver targets exceed 1 from pass 3 on and stall 6.7 and 2.1 cells from the nearest answer.
\emph{Displacement.} At pass 8 the fixed targets still ask for 13.6 (standard) and 8.7 (solver's first answer) changed cells, against 0.18 for ours, as \cref{prop:disp} predicts.
The fixed-solver model solves 83\% of puzzles with 2--4 answers but 35\% of those with 17--64, and its loss stays at 32.7 bits: being learnable in principle did not make its target learned.
Ours solves 97.6--99.5\% in every bin.

\subsection{What matters: the target depends on the input}
\label{sec:res-target}

\begin{figure}[t]
  \centering
  \includegraphics[width=\columnwidth]{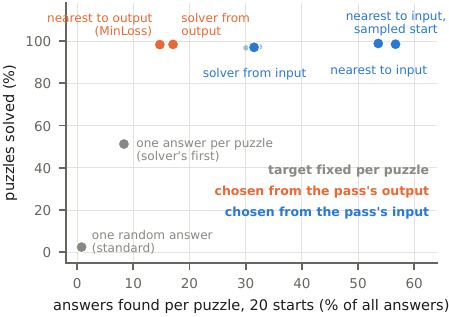}
  \caption{\textbf{What matters is that the target depends on the pass's input.}
  Each point is a target rule (large dot: mean over seeds; small dots: seeds); colour shows what the target is chosen from.}
  \label{fig:target}
\end{figure}

\Cref{fig:target} compares seven target rules, grouped by what the target is chosen from.
Targets fixed per puzzle blend or solve half the puzzles; targets chosen from the pass's \emph{output} solve but find few answers (14.7--17.1\%); targets chosen from the \emph{input} solve and find the most (31.5--56.7\%).
A fixed-point test explains the difference (\cref{lem:fixed}): given a correct answer as its draft, one pass of the nearest-answer model returns it unchanged for 100.0\% of answers, against under 1\% for targets chosen from the output or fixed per puzzle.
Those models solve, but re-route every draft toward a preferred answer.

\subsection{Inside the loop: a weak repairer, iterated}
\label{sec:res-loop}

Each pass fixes far more cells than it breaks (33.8 vs.\ 1.3 at pass 1); wrong cells fall from 45.2 at the start to 10.4, 3.7, 1.4 and 0.66 after passes 1--4.
The answer is chosen early (the nearest answer changes in 61\%, 24\%, 8\% and 3\% of runs after passes 1--4) and then kept, as \cref{lem:stay} predicts (\cref{app:figs}).

\subsection{The target needs the full answer set}
\label{sec:res-lists}

Keeping only $m$ randomly chosen answers per training puzzle shows why a list must be complete.
With $m=1$ the target is fixed and blends (\sd{6.1}{2.2}\%); with $m=2$ and $m=5$ the nearest listed answer solves \sd{27.7}{3.0}\% and \sd{75.9}{1.5}\%, because for a draft near an unlisted answer the target depends on which answers were kept.
A rules checker repairs most of this: the \emph{verified} target, the draft itself if it obeys the rules and else the nearest listed answer, solves \sd{77.0}{2.5}\% with $m=2$, and on the benchmark of \citeauthor{nandwani2021oneofmany}, whose lists are incomplete, it raises solving from \sd{89.3}{1.1}\% to \sd{99.0}{0.1}\%.
The solver target needs no list at all.

\subsection{Controlling quality with a knob}
\label{sec:res-knob}

\begin{figure*}[t]
  \centering
  \includegraphics[width=0.76\textwidth]{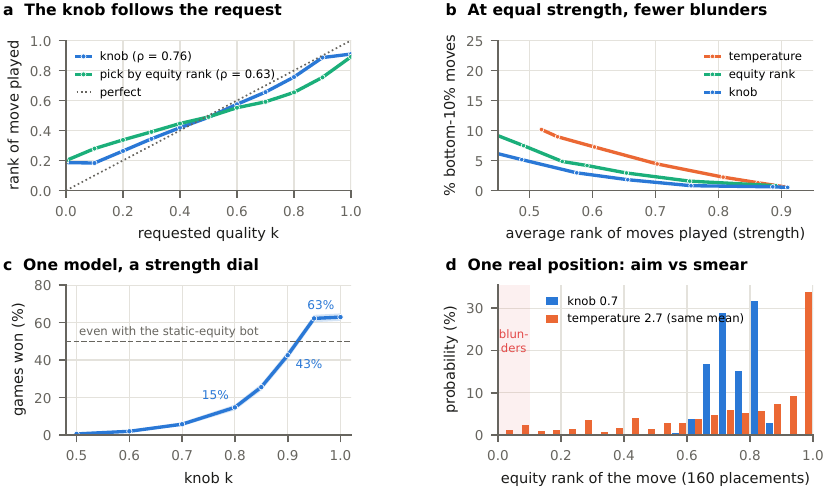}
  \caption{\textbf{A knob turns the target into a strength dial} (Scrabble, $11{\times}11$, 3 seeds).
  (a)~Requested $k$ against the rank of the move played: the knob follows from $k\approx0.2$ up.
  (b)~At equal average strength, the knob plays far fewer bottom-10\% moves than temperature.
  (c)~Games won against MAGPIE's static-equity bot (2,000 per setting; 95\% band).
  (d)~One real position: at $k=0.7$ the knob aims at ranks 0.6--0.85, while temperature of the same mean rank spreads over all ranks, blunders included.
  The knob changes which moves are targets; temperature only adds noise.}
  \label{fig:knob}
\end{figure*}

In Scrabble each legal move is one answer, and quality is its win-probability rank.
The model (2.45M parameters) points at one move and is trained toward the band of moves whose rank lies within 0.1 of a knob $k\sim U[0,1]$.
\Cref{fig:knob} compares the knob with temperature on a no-knob model of the same architecture and with picking the move of nearest static-equity rank.
The knob follows the request ($k=0.3,0.5,0.7,0.9$ give ranks 0.35, 0.49, 0.66, 0.89; Spearman 0.76, against 0.63 and 0.51), and at matched average strength it plays 3.0\% bottom-decile moves against 9.1\% for temperature, while still playing 13--14 different moves in 20 samples.
Against MAGPIE's static-equity bot it wins 5.8\%, 14.7\%, 42.5\% and 62.9\% of games at $k=0.7,0.8,0.9,1.0$.
On Sudoku with a random quality ranking the knob is learned in 2 of 3 seeds (Spearman 0.76 and 0.75).

\subsection{Ablations}
\label{sec:ablations}

On Sudoku (3 seeds), separate weights per pass solve \sd{82.9}{2.6}\% at the same width and \sd{23.0}{15.1}\% at matched parameters; removing the scratchpad gives \sd{90.5}{0.4}\%; and training on expert repair trajectories instead of the model's own drafts gives \sd{74.1}{1.7}\%, as the DAgger view predicts.
Sampling the draft at temperature 1.5 in passes 1--4 raises answers found from 53.9\% to 59.6\% with no loss in solving (\cref{app:weights}).
Adding a ConsFormer-style rule penalty changes nothing measurable (\sd{98.7}{0.3}\%): the proximal target already carries that information.

%=====================================================================
\section{Related work}
\label{sec:related}

\paragraph{Looped and recursive reasoners.}
HRM~\citep{wang2025hrm} and TRM~\citep{jolicoeur2025trm} train one small network, applied repeatedly with deep supervision, toward one answer per puzzle; they are our architecture, and their target is what we replace.
Work on universal and looped transformers and recurrent networks~\citep{dehghani2019universal,saunshi2025looped,merrill2025depth,palm2018rrn,schwarzschild2021learn} studies what looping buys; we study what each pass is trained toward.

\paragraph{Learning with many correct answers.}
\citet{nandwani2021oneofmany} choose the target by the model's prediction (MinLoss) or by a learned selector (SelectR); this solves but collapses onto a few answers (\cref{sec:res-target}).
Multiple-choice learning~\citep{guzman2012mcl,rupprecht2017wta} and set matching~\citep{carion2020detr} match predictions to the closest answer but need a fixed set of output slots, and non-autoregressive translation~\citep{gu2018nat,lee2018iterative,ghazvininejad2019maskpredict,huang2022nat} faces the same mixing of modes.
GRAM~\citep{baek2026gram} adds stochastic latents trained by variational inference; PTRM~\citep{sghaier2026ptrm} adds latent noise at test time and selects among rollouts with the Q head; other work changes how recursive runs are sampled~\citep{corbett2026stochastic,cameron2026drm,carballo2026recursive}.
These methods change \emph{how} a run is drawn; the proximal target changes \emph{what} each pass learns, so that every answer is a fixed point of a deterministic loop.

\paragraph{Iterative refinement without answers.}
ConsFormer~\citep{xu2025consformer} minimises differentiable rule penalties and iterates one step thousands of times.
Diffusion and energy-based reasoners~\citep{ye2025beyond,du2024ired,wewer2025srm} also refine full assignments; flow matching with optimal-transport couplings~\citep{tong2024cfm,pooladian2023multisample} pairs noise with nearby data, an analogue of pairing a draft with its nearest answer; GFlowNets~\citep{bengio2021gflownet} sample in proportion to a reward.

\paragraph{Training toward the best correction of one's own state.}
Dynamic oracles in parsing~\citep{goldberg2012dynamic}, optimal completion distillation~\citep{sabour2019ocd} and DAgger~\citep{ross2011dagger} train a model on its own states toward the best continuation from each state.
The proximal target brings this idea to looped refiners with many answers, and adds the fixed-point and progress guarantees that make a short loop converge.

%=====================================================================
\section{Limitations and conclusion}
\label{sec:conclusion}

\paragraph{Limitations.}
Training needs the full answer set or a way to reach it: a complete list, a solver that can start from a draft, or a ranking; a random sample of answers brings blending back, and a rules checker recovers most but not all of it.
The problems are small (puzzles, an $11{\times}11$ game, models of at most 6.9M parameters), and on GRAM's 10-node colouring benchmark GRAM finds slightly more colourings.
The loop finds 53.6\% of a Sudoku puzzle's answers, below its target's 64.3\% from the same starts; the Scrabble knob is useful for $k\ge0.6$, and the Sudoku knob was learned in 2 of 3 seeds.
No target creates answers of a kind never seen in training: on $10{\times}10$ N-Queens our model solves 99.9\% of puzzles from seen solution families but 2.7\% from held-out ones (\cref{app:full}).

\paragraph{Conclusion.}
When a problem has many correct answers, the per-pass target decides whether a looped refiner can learn at all.
Choosing that target from the pass's own input removes the blending floor, makes every answer a fixed point, and turns training loss into guaranteed progress, with no extra networks and no rules as a formula.
We expect the choice of per-pass target to matter wherever iterative models are trained with deep supervision on outputs that are not unique.

\section*{Impact Statement}
This paper studies how to train small iterative models on combinatorial problems with many correct answers.
Its applications are puzzle solving, constraint satisfaction and strength-adjustable game opponents; we do not foresee negative societal consequences beyond those of machine learning in general.

\bibliography{refs}
\bibliographystyle{icml2026}

%=====================================================================
\newpage
\appendix
\onecolumn

\section{Proofs}
\label{app:proofs}

\paragraph{\Cref{lem:blend} (blending).}
For a fixed input, the expected per-cell cross-entropy $\mathbb{E}[-\log p_i(y_i)\mid \text{input}]$ is minimised by the conditional distribution of the target cell given the input, $p_i^\star(v)=\Pr[y_i=v\mid\text{input}]$.
The input is the puzzle $x$, the draft $s$ and the scratchpad.
By assumption the target is independent of the draft and the scratchpad given $x$, so the conditional equals $\Pr[y_i=v\mid x]$ and does not depend on $s$.
If two possible targets with positive probability disagree in cell $i$, $p_i^\star$ puts positive mass on two values, so $p^\star$ is not a point mass on any correct answer.
For the floor, Gibbs' inequality gives $\mathbb{E}_{y\sim q_x}[-\log p_i(y_i)]\ge H_{q_x}(y_i)$ for any distribution $p_i$, with equality at $p_i=p_i^\star$; summing over free cells gives $F(x)$ for any pass and any capacity, since the bound holds for every function of the input. \hfill$\square$

\paragraph{\Cref{lem:fixed} (fixed points).}
Write $f_s(y)=d(s,y)+c(y)$ and let $y=T_c(s)$.
For every $y'\in S(x)$, optimality of $y$ and the triangle inequality give
$d(s,y)+c(y)\le d(s,y')+c(y')\le d(s,y)+d(y,y')+c(y')$, hence $c(y)\le d(y,y')+c(y')$, that is, $f_y(y)\le f_y(y')$.
So $y$ minimises $f_y$.
If some $y'$ ties with $y$ at the draft $y$, then $c(y')+d(y,y')=c(y)$, and the chain above gives $f_s(y')\le f_s(y)$, so $y'$ was also tied at $s$; the fixed tie order preferred $y$ there and does so again. Hence $T_c(y)=y$.
The same chain shows that an answer $y$ is a fixed point exactly when $c(y)< c(y')+d(y,y')$ for every $y'$ earlier in the tie order and $c(y)\le c(y')+d(y,y')$ for every later $y'$.
With $c=0$ this always holds, since $d(y,y')>0$ for $y'\ne y$.
With $c\in\{0,+\infty\}$ an answer with $c=+\infty$ is never a minimiser, and for answers with $c=0$ the condition holds; so the fixed points are exactly the answers with $c=0$.
Consequently the function that outputs $T_c(s)$ as a point mass attains zero loss in~\eqref{eq:loss} at every pass, and a loop that follows its target from $s_0$ ends at $y$ exactly when $s_0\in T_c^{-1}(y)$, which has probability $\mu(T_c^{-1}(y))$. \hfill$\square$

\paragraph{\Cref{lem:stay} (the target stays put during repair).}
Let $y=T_c(s)$ and let $s'$ set $k$ cells, at which $s$ differs from $y$, to $y$'s values.
Then $d(s',y)=d(s,y)-k$.
Each changed cell changes the distance to any other answer $y'$ by at most one, so $d(s',y')\ge d(s,y')-k$.
Hence $f_{s'}(y)=f_s(y)-k\le f_s(y')-k\le f_{s'}(y')$: the objective drops by exactly $k$ for $y$ and by at most $k$ for every other answer, so $y$ still minimises.
Any $y'$ tied with $y$ at $s'$ must satisfy $f_s(y')=f_s(y)$, so it was tied at $s$ and the tie order again prefers $y$. \hfill$\square$

\paragraph{The solver target.}
Let $\rho(s)$ be the first answer found by depth-first search that picks the most constrained cell first (given the values assigned so far) and tries the draft's value first, then the other values in a fixed order.
The search is deterministic given $s$.
If $s$ is a correct answer, the draft's value is consistent in every cell, so the search never backtracks and returns $s$.
Now let $y=\rho(s)$ and let $s'$ set some cells to $y$'s values.
Induct along the search path that leads to $y$; the cell chosen at each node depends only on the values assigned so far, so the path visits the same cells under $s'$.
At each node, the branches tried before $y$'s value under $s'$ are a subset of those tried before it under $s$: in a repaired cell $y$'s value is now tried first, and in any other cell the order is unchanged.
Under $s$ every earlier branch failed, so its subtree holds no correct answer; these branches fail under $s'$ as well, and the search follows the same path and returns $y$. \hfill$\square$

\paragraph{\Cref{prop:progress} (progress without solving).}
Let $y=T_c(s)$. Since $V$ is a minimum over answers, $V(s')\le d(s',y)+c(y)=e+c(y)$.
For the second claim, a free cell decoded to a value other than $y_i$ has $p_i(y_i)\le 1/2$ under argmax decoding (the decoded value has at least as much mass), so $-\log_2 p_i(y_i)\ge1$; cells decoded to $y_i$ contribute a non-negative amount. Summing over free cells gives $e\le\sum_i-\log_2 p_i(y_i)$ on every draft.
If $e\le(1-\gamma)V(s)$ at every pass (with $c=0$), then $V(s_t)\le(1-\gamma)^tV(s_0)$ by induction. \hfill$\square$

\paragraph{\Cref{prop:disp} (what a fixed target demands).}
If the input is a correct answer $y\ne\kappa(x)$, the target asks for $d(y,\kappa(x))\ge\delta(x)$ changed cells, while $T_0(y)=y$ asks for none (\cref{lem:fixed}).
Two distinct Sudoku solutions differ in at least 4 cells: a row that contains a differing cell holds the same digits in both solutions, so it contains at least two differing cells, and likewise for columns; the differing cells therefore span at least two rows and two columns. \hfill$\square$

\paragraph{Exact checks.}
On $4{\times}4$ Sudoku every draft can be enumerated. We used 60 puzzles with 6--10 free cells (20 each with 2, 3--4 and 5--6 answers) and the nearest-answer target with the fixed tie order.
\emph{Exhaustive:} on all 40.3M drafts, $T_0(T_0(s))=T_0(s)$, every answer is a fixed point, and each of the 250.5M single repairs toward the target keeps the target (0 violations). Single repairs suffice for \cref{lem:stay}, since any set of repairs is a chain of single ones.
\emph{Sampled:} the fixed-point condition of \cref{lem:fixed} with random integer costs (0 mismatches); \cref{prop:progress} on 5,000 random pairs of drafts per puzzle (0 violations); multi-cell repairs with ties broken at random, on 3,000 sampled drafts per puzzle with up to 256 repair subsets each (14.8M repairs; 1.8\%, 2.5\% and 3.3\% violations for the three groups, so the fixed tie order is needed); and the minimiser of \cref{lem:blend} by gradient descent, which matches the marginals to within $5\cdot10^{-4}$.
In the trained models (\cref{sec:res-theory}), $V(s_{t+1})\le e_t\le$ loss in bits held for every run of every model.

%---------------------------------------------------------------------
\section{Experimental details}
\label{app:details}

\paragraph{Sudoku (2--64 answers).}
Puzzles are generated from random complete grids, with clues removed while the answer stays unique, down to 26--40 clues.
For half of the puzzles more clues are removed until there are 2--64 correct answers (the target count is drawn log-uniformly), and an exact solver enumerates all answers.
We use 500,000 training, 2,000 validation and 4,000 test puzzles, generated with disjoint seeds.
Multi-answer test puzzles have 53 free cells and 22.6 answers on average (median 16); two answers differ in 16.6 cells on average.
No test solution grid equals a training solution grid, even after relabelling digits, and a lookup baseline that returns the nearest training solution consistent with the clues solves 0\%.

\paragraph{Sparse Sudoku.}
17--28 clues are kept from a random grid; 500k / 2k / 4k puzzles. Validity is checked by the rules, and training targets come from the solver.

\paragraph{Planted 3-colouring.}
Random graphs with a planted balanced 3-colouring and mean degree 4; 15\% of nodes are pre-coloured. Training graphs have 20--50 nodes and are generated on the fly; test sets have 1,000 graphs each of 30, 50, 100 and 200 nodes.
The refiner has no positional embedding, so it is equivariant to node relabelling, and attention is restricted to graph neighbours.

\paragraph{Benchmark of \citet{nandwani2021oneofmany}.}
The released files are converted without modification; for each training puzzle we keep at most 5 listed answers and use digit-relabelling augmentation.

\paragraph{N-Queens audit.}
Partial placements with 7, 8 or 9 queens on a $10{\times}10$ board are completed to 10 non-attacking queens. We hold out 20\% of the solution families (symmetry orbits of the 724 solutions under rotations and reflections); test puzzles whose solutions all lie in held-out families are ``new''.

\paragraph{Scrabble.}
An $11{\times}11$ board with the standard tile distribution; 45,000 / 2,500 / 2,500 positions from separate self-play games.
Every legal placement (at most 512 per position) is labelled with its win probability by 2-ply Monte Carlo simulation in MAGPIE, and its quality rank is its percentile among the position's legal moves (1 is best).
The model encodes every legal placement as a token, attends over board, rack and candidate tokens with the same looped architecture (4 passes), and points at one candidate; the draft is not used.
The knob enters as a zero-initialised projection of Fourier features of $k$, added to every token; band width $w=0.2$; the loss is $-\log\sum_{|r(y)-k|\le w/2} p(y)$.
Baseline dials: temperature on the same architecture trained without a knob toward the soft win-probability distribution; and the move whose static-equity rank is nearest $k$.
Each setting plays 2,000 games against MAGPIE's static-equity bot (1,000 bag seeds, both seat orders).

\paragraph{Puzzle model and training.}
One token per cell, embedding the clue, the draft value and the position; two transformer blocks of width 256 with 8 heads; two latent states updated in the TRM pattern (4 lower updates and 1 upper update per pass). The draft embedding is zero-initialised.
AdamW with learning rate $3\cdot10^{-4}$, warm-up over 20k examples, weight decay 0.1, $\beta=(0.9,0.95)$, gradient clipping at 1.0, batch 256, and an exponential moving average of the weights.
Each run sees 3M training examples; we validate every 200k examples and test the best snapshot once.
Deep supervision uses one backward pass per pass, with the latent detached between passes.
Each run uses one RTX 2080 Ti.

\paragraph{ConsFormer (reimplemented from \citet{xu2025consformer}).}
An encoder-only, post-norm transformer with 7 layers (Sudoku) or 4 (colouring), width 128 and 4 heads (a variant with the paper's 3 heads at width 126 gives the same results); 1.84M and 1.05M parameters.
It is trained self-supervised from a random assignment, with a random subset of cells updated through Gumbel-softmax, on the sum of squared relaxed rule violations, for 20M examples; at test time it is iterated $T\in\{8,100,1{,}000,2{,}000\}$ times.
The 8-pass variant unrolls 8 iterations per update and averages their penalties.

\paragraph{Measures.}
Answers found: over 20 random starts, the share of the puzzle's correct answers reached. Twenty independent uniform draws over each puzzle's answers would find 68.9\%; the nearest-answer target itself, applied to the same uniform starts, 64.3\%.
Fixed point: the share of correct answers that one pass returns unchanged when given as the draft with a fresh scratchpad.
Theory measures: $V$, the pass's errors $e_t$ against its own training target and its loss in bits, at every pass, on 1,000 multi-answer puzzles $\times$ 4 starts.

%---------------------------------------------------------------------
\section{GRAM: protocol and deviations}
\label{app:gram}

We ran the official GRAM code (MIT licence) without changing its model or training logic.

\paragraph{Reproduction on its own benchmark.}
GRAM's 10-node graph colouring data and its published configuration (attention, $H=3$, $L=4$, KL 0.45, learning rate $10^{-4}$, weight decay 1.0, EMA), 3 seeds, 192 test graphs, 20 samples, measured with its own definition of Cov@20 (colourings in canonical form).
\emph{Deviations, forced by hardware:} one RTX 2080 Ti instead of 8 RTX 4090, so batch 256 instead of 768, float32 instead of bfloat16, and no \texttt{torch.compile}.
Without compilation its posterior one-hot-encodes labels that hold the ignore index; we map ignored positions to its padding class through an import hook, leaving its code untouched.
At its published number of optimiser steps the seed mean was 45.9\%; continuing to 3,400 epochs (about 46M examples, against its published 67M) gives \sd{50.0}{2.8}\%, within one standard deviation of the published \sd{51.3}{2.8}\%, so we use these models.
Our solver-target refiner and a standard-target refiner were trained on the same training graphs (10\% held out for validation) and measured with the same code after canonicalising colours, since our refiner has no notion of a canonical colour order; GRAM's score barely changes under canonicalisation (50.0\% before, 50.2\% after).
Single-sample conflict edges, summed over the 192 test graphs: GRAM 15.0, ours 27.3, standard target 216.0.

\paragraph{Sudoku.}
Our Sudoku with 2--64 answers, converted to its format, with one (puzzle, answer) training example per listed answer, so GRAM trains on every answer (5.9M pairs).
Its evaluation split is our validation split; our test set stays unseen.
\emph{Configuration selection} (seed 0, validation, one epoch each): its Sudoku configuration (MLP mixer, no positional encoding, $L=6$) and its multi-solution configuration (attention, rotary encoding, $L=4$), each with KL $\in\{0.045, 0.1, 0.45\}$.
The MLP configuration with KL 0.045 had the highest validation solved (59.0\%) and answers found (37.9\%); with KL 0.1 it was within one point on solved (58.1\%), so both received 3 seeds.
\emph{Budget:} 4 epochs (23.6M examples), equal to our 3M examples times 8 supervised passes. Its validation accuracy did not rise after the first epoch (0.53--0.64 at every epoch), so we did not train longer. Batch 64 (memory), float32.
\emph{Results} (\cref{tab:gram}): GRAM is evaluated with its own 16 adaptive steps and 20 samples.
We did not run GRAM on our planted colouring: its encoding is the flattened adjacency matrix, $n(n{+}1)/2$ tokens, which is 1,275 tokens at 50 nodes and 20,100 at 200, beyond our hardware.

\begin{table}[h]
  \caption{\textbf{GRAM on Sudoku with 2--64 answers} (test set, 20 samples, mean $\pm$ sd over 3 seeds). Q-selected: the sample chosen by GRAM's Q head is correct. Distinct: different valid answers among 20 samples.}
  \label{tab:gram}
  \centering
  \small
  \setlength{\tabcolsep}{4pt}
  \begin{tabular}{@{}lccccccc@{}}
    \toprule
    & \multicolumn{6}{c}{Multi-answer puzzles} & Unique \\
    \cmidrule(lr){2-7}
    Model & Solved & Any of 20 & Q-sel. & Found & Found, 2--4 & Found, 17--64 & Solved \\
    \midrule
    GRAM, KL 0.045 & \sd{57.7}{5.7} & \sd{96.6}{0.4} & \sd{65.7}{7.4} & \sd{37.4}{1.9} & 70.6 & 20.8 & 64.5 \\
    GRAM, KL 0.1 & \sd{44.6}{16.1} & \sd{67.3}{11.0} & \sd{51.6}{15.9} & \sd{12.6}{1.7} & 39.5 & 3.1 & 88.9 \\
    Proximal (nearest answer) & \sd{98.8}{0.3} & -- & -- & \sd{53.6}{0.4} & -- & -- & \sd{98.6}{0.3} \\
    \bottomrule
  \end{tabular}
\end{table}

%---------------------------------------------------------------------
\section{Full results}
\label{app:full}
All entries are mean $\pm$ sd over 3 seeds unless stated otherwise; solved and answers found are in \%.

\begin{table}[h]
  \caption{\textbf{Sudoku with 2--64 answers: all variants.} Mean $\pm$ sd over 5 seeds (first two rows) or 3. Fixed pt.: one pass with a fresh scratchpad returns a correct draft unchanged. The solver-target model fails that test, and starting 10\% of its training runs at a correct answer fixes it.}
  \label{tab:full-sudoku}
  \centering
  \small
  \begin{tabular}{@{}lcccc@{}}
    \toprule
    Variant & Solved (multi) & Solved (unique) & Answers found & Fixed pt. \\
    \midrule
    standard target (blank start) & \sd{2.5}{0.2} & \sd{96.3}{0.4} & \sd{0.8}{0.0} & \sd{0.1}{0.1} \\
    nearest to input (ours) & \sd{98.8}{0.3} & \sd{98.6}{0.3} & \sd{53.6}{0.4} & \sd{100.0}{0.0} \\
    \midrule
    standard, random start & \sd{4.3}{2.9} & \sd{94.1}{3.9} & \sd{1.7}{1.2} & \sd{0.1}{0.1} \\
    standard, new random answer every pass & \sd{3.2}{1.5} & \sd{95.5}{1.5} & \sd{1.3}{0.4} & \sd{0.1}{0.0} \\
    standard, 4$\times$ parameters (6.9M) & \sd{2.4}{0.1} & \sd{96.6}{0.2} & \sd{0.8}{0.0} & \sd{0.1}{0.0} \\
    solver's first answer, fixed per puzzle & \sd{51.3}{0.7} & \sd{96.9}{0.2} & \sd{8.3}{0.1} & \sd{0.3}{0.1} \\
    nearest to output (MinLoss) & \sd{98.3}{0.3} & \sd{97.8}{0.1} & \sd{14.7}{0.2} & \sd{0.4}{0.1} \\
    nearest to input, blank start & \sd{98.8}{0.2} & \sd{98.5}{0.2} & \sd{10.6}{0.1} & \sd{100.0}{0.0} \\
    nearest to input, 4$\times$ parameters & \sd{99.1}{0.1} & \sd{98.8}{0.3} & \sd{55.0}{0.8} & \sd{100.0}{0.0} \\
    nearest to input, sampled start & \sd{98.5}{0.2} & \sd{98.3}{0.1} & \sd{56.7}{0.4} & \sd{100.0}{0.0} \\
    solver started from output & \sd{98.4}{0.4} & \sd{97.8}{0.2} & \sd{17.1}{0.3} & \sd{0.3}{0.1} \\
    solver started from input & \sd{97.0}{0.2} & \sd{98.4}{0.4} & \sd{31.5}{1.3} & \sd{2.9}{0.7} \\
    \quad + 10\% of training starts correct & \sd{97.4}{0.6} & \sd{98.5}{0.3} & \sd{33.3}{1.4} & \sd{99.9}{0.2} \\
    ours, separate weights, same width & \sd{82.9}{2.6} & \sd{91.4}{1.2} & \sd{50.0}{0.3} & \sd{100.0}{0.0} \\
    ours, no scratchpad & \sd{90.5}{0.4} & \sd{93.0}{0.6} & \sd{51.0}{0.4} & \sd{100.0}{0.0} \\
    separate weights, no scratchpad & \sd{88.6}{0.7} & \sd{91.0}{0.3} & \sd{52.4}{0.4} & \sd{100.0}{0.0} \\
    separate weights, matched params & \sd{23.0}{15.1} & \sd{67.9}{7.8} & \sd{19.3}{8.8} & \sd{98.3}{1.8} \\
    ours, expert-trajectory training & \sd{74.1}{1.7} & \sd{82.3}{0.3} & \sd{49.4}{0.1} & \sd{100.0}{0.0} \\
    ours + rule penalty, $w=1$ & \sd{98.7}{0.3} & \sd{98.5}{0.2} & \sd{53.2}{0.4} & \sd{100.0}{0.0} \\
    ours + rule penalty, $w=10$ & \sd{98.6}{0.2} & \sd{98.4}{0.2} & \sd{51.4}{0.5} & \sd{100.0}{0.0} \\
    \bottomrule
  \end{tabular}
\end{table}

\begin{table}[h]
  \caption{\textbf{Incomplete answer lists bring blending back; a rules checker recovers most of it.} $m$ randomly chosen answers kept per training puzzle (all answers count at test time). Verified: the draft itself if it obeys the rules, else the nearest listed answer. Unlisted fixed pt.: answers never used as targets.}
  \label{tab:lists}
  \centering
  \small
  \begin{tabular}{@{}lcccc@{}}
    \toprule
    Target & Solved (multi) & Answers found & Fixed pt. & Unlisted fixed pt. \\
    \midrule
    nearest listed, $m=1$ & \sd{6.1}{2.2} & \sd{2.5}{0.8} & \sd{0.0}{0.0} & -- \\
    nearest listed, $m=2$ & \sd{27.7}{3.0} & \sd{9.1}{0.8} & \sd{0.8}{0.0} & \sd{0.1}{0.1} \\
    nearest listed, $m=5$ & \sd{75.9}{1.5} & \sd{30.0}{4.4} & \sd{49.8}{43.6} & \sd{44.8}{47.3} \\
    verified, $m=2$ & \sd{77.0}{2.5} & \sd{21.8}{0.7} & \sd{16.6}{2.1} & \sd{10.6}{2.3} \\
    all answers listed (reference) & \sd{98.8}{0.3} & \sd{53.6}{0.4} & \sd{100.0}{0.0} & -- \\
    \bottomrule
  \end{tabular}
\end{table}

\begin{table}[h]
  \caption{\textbf{Measured theory quantities} (1,000 multi-answer puzzles $\times$ 4 starts, mean of 3 seeds). Ratio: the pass's errors against its own target over the input's distance $V$ to the nearest answer (below 1 guarantees progress). Loss: bits per puzzle at pass 8 (floor of \cref{lem:blend}: 35.5). Asked: $d(s_t,y_t)$ at pass 8. $V_8$: distance to the nearest answer before the last pass. Warm: keeps its answer within a run.}
  \label{tab:theory}
  \centering
  \small
  \begin{tabular}{@{}lcccccc@{}}
    \toprule
    Target & Ratio, pass 1 / 4 & Loss & Asked & $V_8$ & Fixed pt. & Warm \\
    \midrule
    standard & 0.30 / 2.29 & 43.4 & 13.57 & 6.68 & 0.001 & 0.87 \\
    standard, 4$\times$ parameters & 0.29 / 2.41 & 42.7 & 13.41 & 6.43 & 0.000 & 0.92 \\
    solver's first answer (fixed) & 0.32 / 2.95 & 32.7 & 8.65 & 2.08 & 0.002 & 0.93 \\
    nearest to output (MinLoss) & 0.27 / 0.41 & 1.0 & 0.36 & 0.33 & 0.004 & 1.00 \\
    solver from output & 0.27 / 0.39 & 0.8 & 0.28 & 0.25 & 0.003 & 1.00 \\
    solver from input & 0.33 / 0.97 & 1.6 & 0.49 & 0.38 & 0.026 & 1.00 \\
    nearest to input (ours) & 0.28 / 0.35 & 0.5 & 0.18 & 0.18 & 1.000 & 1.00 \\
    \bottomrule
  \end{tabular}
\end{table}

\begin{table}[h]
  \caption{\textbf{Sparse Sudoku (more than 10,000 answers): the solver started from the input gives the most distinct answers.} Distinct: different correct answers among 20 starts. Fixed pt.\ (fresh): the fixed-point test on answers never used as targets.}
  \label{tab:full-sparse}
  \centering
  \small
  \begin{tabular}{@{}lccc@{}}
    \toprule
    Variant & Solved & Distinct (of 20) & Fixed pt. (fresh) \\
    \midrule
    standard target (5 seeds) & \sd{0.0}{0.0} & \sd{0.0}{0.0} & \sd{0.0}{0.0} \\
    solver's first answer, fixed per puzzle & \sd{37.3}{4.7} & \sd{0.6}{0.1} & \sd{0.0}{0.0} \\
    solver started from output & \sd{99.4}{0.2} & \sd{10.0}{0.4} & \sd{0.0}{0.0} \\
    solver started from input (ours; 5 seeds) & \sd{93.8}{0.7} & \sd{15.1}{0.1} & \sd{15.7}{1.7} \\
    \bottomrule
  \end{tabular}
\end{table}

\begin{table}[h]
  \caption{\textbf{Planted 3-colouring: the solver target generalises from 20--50 to 200 nodes.} Solved by graph size $n$ (8 passes; ``$n{=}200$, 64'': 64 passes at test time), and distinct colourings among 20 starts at $n=200$.}
  \label{tab:full-color}
  \centering
  \small
  \begin{tabular}{@{}lcccccc@{}}
    \toprule
    Variant & $n{=}30$ & $n{=}50$ & $n{=}100$ & $n{=}200$ & $n{=}200$, 64 & Distinct \\
    \midrule
    standard (5 seeds) & \sd{71.9}{1.5} & \sd{71.9}{3.1} & \sd{57.8}{3.9} & \sd{36.9}{5.6} & \sd{37.7}{5.4} & \sd{0.9}{0.3} \\
    solver from input (ours; 5 seeds) & \sd{99.0}{0.4} & \sd{98.3}{0.4} & \sd{96.5}{0.4} & \sd{90.9}{1.0} & \sd{91.8}{1.1} & \sd{9.2}{0.1} \\
    ours, separate weights & \sd{97.7}{0.4} & \sd{98.4}{0.4} & \sd{95.9}{0.6} & \sd{93.5}{0.5} & \sd{89.6}{6.4} & \sd{8.9}{0.5} \\
    rule penalty only (no target) & \sd{79.1}{3.4} & \sd{75.5}{5.1} & \sd{54.9}{8.4} & \sd{31.6}{7.2} & \sd{35.5}{9.9} & \sd{3.2}{1.2} \\
    no loop, 10 blocks (8.5M) & \sd{48.9}{0.9} & \sd{45.0}{1.4} & \sd{16.8}{1.2} & \sd{3.1}{0.2} & \sd{3.1}{0.2} & \sd{0.3}{0.0} \\
    \bottomrule
  \end{tabular}
\end{table}

\begin{table}[h]
  \caption{\textbf{Harder colouring families: the solver target works everywhere; the standard target needs clues.} 100-node planted graphs by mean degree and pre-coloured share (models trained at degree 4, 15\%); 3 seeds, 500 graphs each. Distinct: valid colourings among 10 starts, counted up to relabelling of colours when nothing is pre-coloured.}
  \label{tab:families}
  \centering
  \small
  \begin{tabular}{@{}cccccc@{}}
    \toprule
    Mean degree & Pre-coloured & Standard, 8 passes & Ours, 8 passes & Ours, 32 passes & Distinct (std / ours) \\
    \midrule
    3.5 & 0\% & \sd{0.0}{0.0} & \sd{84.8}{7.2} & \sd{87.5}{3.5} & 0.0 / 8.8 \\
    3.5 & 5\% & \sd{2.2}{0.3} & \sd{84.9}{1.8} & \sd{86.1}{2.2} & 0.0 / 8.7 \\
    3.5 & 15\% & \sd{35.0}{0.9} & \sd{91.9}{1.1} & \sd{92.0}{2.3} & 0.6 / 9.2 \\
    4.0 & 0\% & \sd{0.0}{0.0} & \sd{79.4}{16.6} & \sd{93.0}{5.1} & 0.0 / 9.3 \\
    4.0 & 5\% & \sd{15.4}{0.5} & \sd{87.3}{8.1} & \sd{93.8}{3.1} & 0.4 / 9.4 \\
    4.0 & 15\% & \sd{58.3}{1.5} & \sd{95.8}{0.7} & \sd{96.2}{1.0} & 0.9 / 9.7 \\
    4.3 & 0\% & \sd{0.0}{0.0} & \sd{61.9}{17.8} & \sd{90.5}{9.6} & 0.0 / 9.0 \\
    4.3 & 5\% & \sd{34.8}{2.6} & \sd{81.3}{8.1} & \sd{93.3}{5.6} & 0.9 / 9.4 \\
    4.3 & 15\% & \sd{71.1}{1.9} & \sd{97.5}{0.8} & \sd{98.0}{1.1} & 1.1 / 9.8 \\
    4.6 & 0\% & \sd{0.1}{0.1} & \sd{41.9}{12.6} & \sd{85.5}{9.2} & 0.0 / 8.5 \\
    4.6 & 5\% & \sd{46.8}{1.1} & \sd{78.4}{5.3} & \sd{94.3}{3.3} & 1.2 / 9.4 \\
    4.6 & 15\% & \sd{78.2}{0.6} & \sd{98.1}{0.8} & \sd{98.2}{0.9} & 1.0 / 9.8 \\
    \bottomrule
  \end{tabular}
\end{table}

\begin{table}[h]
  \caption{\textbf{Left: the benchmark of \citet{nandwani2021oneofmany}; the verified target removes most of the loss from its incomplete lists. Right: the N-Queens audit; no method transfers to new solution families.} Found: answers found by 20 starts.}
  \label{tab:full-nandwani-nq}
  \centering
  \small
  \begin{tabular}{@{}lccc@{}}
    \toprule
    Variant & Solved (multi) & Found & Solved (unique) \\
    \midrule
    standard target (5 seeds) & \sd{6.3}{4.0} & \sd{2.3}{1.0} & \sd{96.7}{3.4} \\
    nearest to input (5 seeds) & \sd{89.3}{1.1} & \sd{49.4}{1.4} & \sd{99.0}{0.2} \\
    verified target & \sd{99.0}{0.1} & \sd{52.2}{1.4} & \sd{98.9}{0.1} \\
    \bottomrule
  \end{tabular}\hspace{2em}
  \begin{tabular}{@{}lcccc@{}}
    \toprule
    & \multicolumn{2}{c}{Seen families} & \multicolumn{2}{c}{New families} \\
    \cmidrule(lr){2-3}\cmidrule(lr){4-5}
    Variant & Solved & Found & Solved & Found \\
    \midrule
    ours & \sd{99.9}{0.1} & \sd{98.7}{0.1} & \sd{2.7}{0.9} & \sd{7.0}{2.9} \\
    MinLoss & \sd{96.3}{5.1} & \sd{42.7}{0.9} & \sd{11.0}{5.1} & \sd{16.9}{2.7} \\
    \bottomrule
  \end{tabular}
\end{table}

\begin{table}[h]
  \caption{\textbf{ConsFormer needs hundreds to thousands of passes.} Solved by passes at test time $T$; answers found on Sudoku at $T=1{,}000$.}
  \label{tab:full-consformer}
  \centering
  \small
  \begin{tabular}{@{}lccccc@{}}
    \toprule
    Sudoku & $T{=}8$ & $T{=}100$ & $T{=}1{,}000$ & $T{=}2{,}000$ & Unique, $T{=}2$k \\
    \midrule
    solved & \sd{0.1}{0.1} & \sd{35.6}{0.6} & \sd{92.5}{0.5} & \sd{97.5}{0.2} & \sd{95.9}{0.5} \\
    answers found & & & \sd{64.3}{0.3} & & \\
    solved, 3 heads & \sd{0.2}{0.1} & \sd{35.4}{0.7} & \sd{91.8}{1.0} & \sd{97.3}{0.5} & \sd{94.4}{1.4} \\
    solved, 8-pass training & \sd{0.0}{0.0} & & & & \\
    \midrule
    Colouring & 30 nodes, $T{=}8$ & 30, $T{=}2$k & 200, $T{=}8$ & 200, $T{=}2$k & \\
    \midrule
    solved & \sd{5.0}{0.2} & \sd{89.3}{1.9} & \sd{0.0}{0.0} & \sd{74.8}{5.0} & \\
    \bottomrule
  \end{tabular}
\end{table}

\paragraph{Knob on Sudoku (3 seeds).}
Answers carry a fixed random quality, ranked within each puzzle; band width 0.25. Spearman correlation between $k$ and the rank reached: 0.763, 0.380 and 0.751 (seeds 0, 1, 2). In seeds 0 and 2, the rank reached at $k=0,0.25,0.5,0.75,1$ is 0.11--0.12, 0.23--0.24, 0.49--0.50, 0.76 and 0.88--0.89, with 72--95\% solved; at $k=0$ and $1$, an answer near $k$ given as the draft is kept 86\% of the time, a far one 40--44\%. Seed 1 did not learn the knob.

\paragraph{Scrabble dials (3 seeds).}
Spearman correlation between setting and rank reached: knob \sd{0.764}{0.002}; equity-rank dial 0.633; temperature 0.513; oracle 0.976.
Requested $k\to$ rank reached / share of bottom-10\% moves (seed 0): 0.0 $\to$ 0.19 / 48.6\%; 0.3 $\to$ 0.35 / 12.0\%; 0.5 $\to$ 0.49 / 4.8\%; 0.7 $\to$ 0.66 / 1.7\%; 0.9 $\to$ 0.89 / 0.7\%; 1.0 $\to$ 0.91 / 0.4\%.
At matched average strength: knob $k=0.6$ blunders 3.0\%, temperature 9.1\%, equity rank 7.5\%.
Games won (2,000 each): $k$ = 0.5: 0.6\%; 0.6: 2.0\%; 0.7: 5.8\%; 0.8: 14.7\%; 0.85: 25.5\%; 0.9: 42.5\%; 0.95: 62.2\%; 1.0: 62.9\%. The no-knob model at temperature 0 wins 60.3\%.

\paragraph{Test-time noise on the standard target (3 seeds).}
Latent noise $\sigma\in\{0, 0.1, 0.3, 0.6, 1.0\}$ at every pass, 20 runs: valid runs 2.3, 2.6, 4.5, 0.8, 0.0\%; puzzles with any valid run 2.3, 8.4, 24.4, 8.3, 0.0\%; answers found 0.8, 2.8, 8.8, 3.3, 0.0\%.

%---------------------------------------------------------------------
\section{Additional figures}
\label{app:figs}

\begin{figure}[h]
  \centering
  \includegraphics[width=0.85\textwidth]{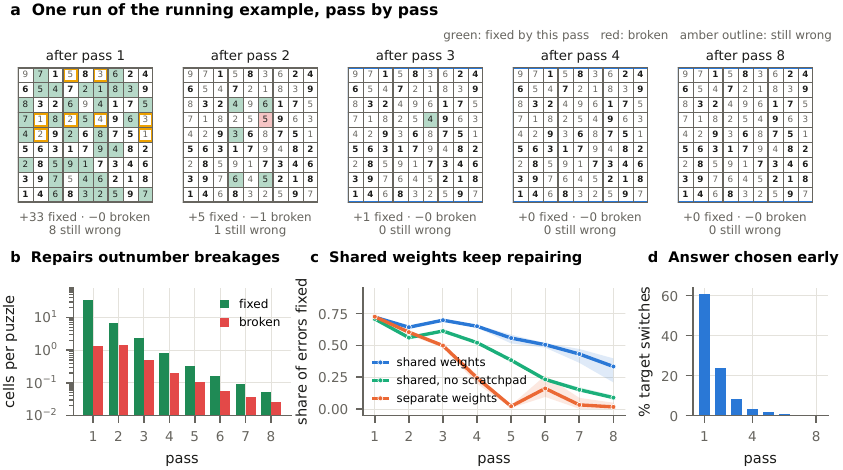}
  \caption{\textbf{The loop is a weak repairer, iterated.}
  (a)~The running example, one start, pass by pass. Green: cells this pass fixed; red: cells it broke; amber outline: still wrong.
  (b)~Test set (3 seeds): cells fixed vs.\ broken per pass (log scale).
  (c)~Net share of the remaining wrong cells fixed per pass: shared weights keep repairing; separate weights per pass stop by pass~5; without the scratchpad, repair fades.
  (d)~The nearest answer switches in 61\%, 24\%, 8\% and 3\% of runs after passes 1--4.}
  \label{fig:loop}
\end{figure}

\begin{figure}[h]
  \centering
  \includegraphics[width=0.85\textwidth]{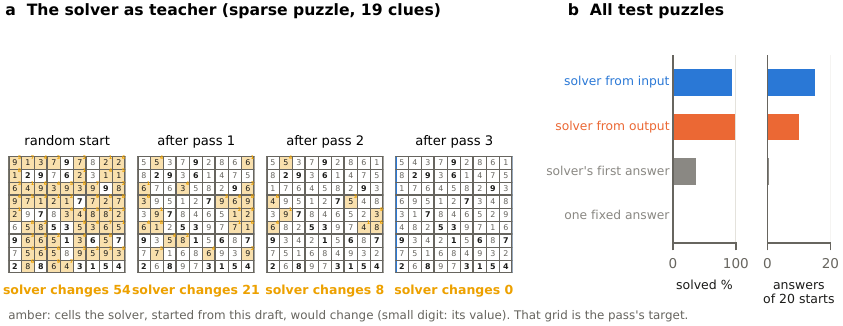}
  \caption{\textbf{A search solver replaces the answer list.}
  (a)~A sparse puzzle (19 clues, more than 10,000 answers). Amber marks the cells that a depth-first solver, started from the current draft, would change; its answer is the target for the next pass. The corrections shrink from 54 cells to 21, 8 and 0.
  (b)~Test set, solved and distinct answers of 20 starts, by target.}
  \label{fig:teacher}
\end{figure}

\begin{figure}[h]
  \centering
  \includegraphics[width=0.7\textwidth]{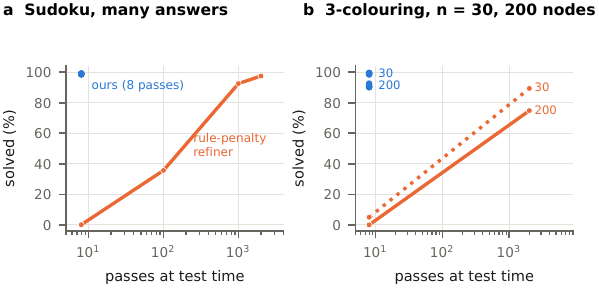}
  \caption{\textbf{A rule-penalty refiner and a short trained loop work in different regimes.}
  Solved against passes at test time: ConsFormer (orange) and ours at 8 passes (blue). (a)~Sudoku; (b)~colouring, 30 and 200 nodes.}
  \label{fig:baselines}
\end{figure}

%---------------------------------------------------------------------
\section{Inside the trained weights; temperature sweep}
\label{app:weights}

\begin{figure}[h]
  \centering
  \includegraphics[width=0.85\textwidth]{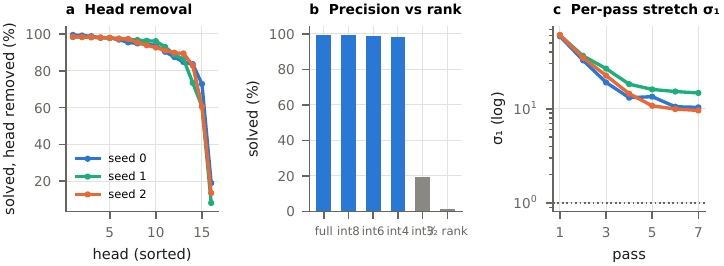}
  \caption{\textbf{The loop converges because its targets are fixed points, not because the map shrinks.}
  Inside the trained weights of the nearest-answer model on Sudoku (3 seeds).
  (a)~Removing one attention head at a time: one head (a different one per seed) is essential (solved drops to 19 / 8 / 14\%); 5 of 16 heads matter (solved below 90\%).
  (b)~8-, 6- and 4-bit weights keep 97.6--99.6\% solved; 3-bit weights and halving every weight matrix's rank break it.
  (c)~The largest stretch $\sigma_1$ of one pass's latent map falls from about 60 at pass~1 to 10--15 at pass~7 but stays far above 1: the loop is not a contraction.}
  \label{fig:weights}
\end{figure}

\paragraph{Weights.}
Figure~\ref{fig:weights} probes the trained weights.
Removing the most important head drops solving to 19.1 / 8.3 / 13.7\% (seeds 0 / 1 / 2).
With int8 / int6 / int4 weights, the seeds solve 99.3 / 99.2 / 98.2\%, 98.6 / 98.8 / 97.7\% and 99.6 / 99.4 / 99.0\%.
With int3 weights they solve 45.4 / 9.7 / 3.5\%, and with halved rank 0.2 / 0.0 / 3.2\%.
The largest singular value $\sigma_1$ of one pass's latent Jacobian is 58.8 / 60.3 / 60.7 at pass~1 and 10.4 / 14.7 / 9.6 at pass~7.

\begin{figure}[h]
  \centering
  \begin{minipage}[c]{0.52\linewidth}
    \includegraphics[width=\linewidth]{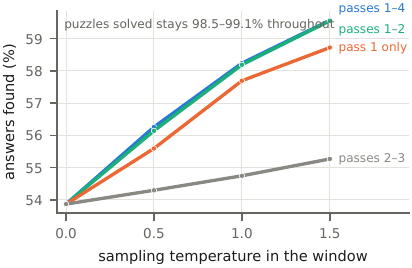}
  \end{minipage}\hfill
  \begin{minipage}[c]{0.45\linewidth}
    \caption{\textbf{Sampling the early passes raises answers found at no cost in validity.}
    The draft is sampled at temperature $\tau$ in a window of passes (3 seeds, mean).
    Sampling in the first passes raises answers found from 53.9\% to 59.6\% ($\tau=1.5$, passes 1--4); sampling only in passes 2--3 helps little.
    Solved stays 98.5--99.1\% throughout.
    Diversity is decided where the answer is chosen: in the first passes.}
    \label{fig:temperature}
  \end{minipage}
\end{figure}

\paragraph{Temperature sweep.}
Figure~\ref{fig:temperature} samples the draft at temperature $\tau$ in a window of passes, at test time only, for the nearest-answer model; it is the source of the temperature result in \cref{sec:ablations}.

\end{document}